\documentclass{article}

\usepackage{PRIMEarxiv}
\usepackage{graphicx}
\usepackage{float}  
\usepackage[utf8]{inputenc} 
\usepackage[T1]{fontenc}    
\usepackage{hyperref}       
\usepackage{url}            
\usepackage{booktabs}       
\usepackage{amsfonts}       
\usepackage{nicefrac}       
\usepackage{microtype}      
\usepackage{lipsum}
\usepackage{graphicx}
\graphicspath{{media/}}     
\usepackage{amsmath}
  
\title{Hand Shape and Gesture Recognition using Multiscale Template Matching, Background Subtraction and Binary Image Analysis
}

\author{
  Ketan Suhaas Saichandran \\
  Graduate School of Arts \& Sciences\\
  Boston University\\
  Boston, MA 02215\\
  \texttt{ketanss@bu.edu} \\
}

\begin{document}
\maketitle

\begin{abstract}
This paper presents a hand shape classification approach employing multiscale template matching. The integration of background subtraction is utilized to derive a binary image of the hand object, enabling the extraction of key features such as centroid and bounding box. The methodology, while simple, demonstrates effectiveness in basic hand shape classification tasks, laying the foundation for potential applications in straightforward human-computer interaction scenarios. Experimental results highlight the system's capability in controlled environments.
\end{abstract}


\section{Introduction}
This study addresses the challenge of hand shape and gesture recognition in computer vision applications, with a focus on contributing to the development of effective human-computer interaction systems. The primary objective is the accurate classification of hand shapes, crucial for applications such as gesture-based interfaces and sign language recognition. Our research aims to enhance the precision and efficiency of existing recognition systems, thereby improving user experiences across various technological domains. Assumptions are made within controlled environments, where stable lighting conditions and clear hand-background distinction are presumed. However, real-world scenarios introduce challenges such as camera exposure issues. To mitigate these challenges, we implemented a strategy to adjust exposure settings and capture a background frame for proper background subtraction. Gaussian blur applications were employed to address noise and disturbances in the environment. Lighting disturbances were considered in the difference image, acknowledging potential errors in centroid calculations. Difficulties extend to the identification and preprocessing of hand shape templates, a critical aspect of our approach. In response, we employed templates for common hand gestures like thumbs up, scissors, paper, and rock. Recognizing the limitations of static hand shapes, we incorporated motion analysis of the centroid to detect hand movement. These solutions aim to enhance the robustness of our system against anticipated difficulties and provide a more accurate hand shape and gesture recognition framework. Importantly, this study explicitly avoids the use of deep learning techniques. This choice is particularly valuable when dealing with limited data, as traditional algorithms demonstrate their efficacy. The presented approach serves as a practical alternative, offering reliable results in scenarios where deep learning models might face challenges due to data scarcity. This research contributes to the broader understanding of hand shape and gesture recognition, emphasizing the applicability and benefits of traditional algorithms in real-world, resource-constrained environments.

\section{Method and Implementation}
This section provides a comprehensive overview of the methodologies and implementation details employed in developing the hand shape detection system. The central emphasis lies on harnessing computer vision techniques, specifically leveraging the powerful capabilities of OpenCV. \cite{OpenCV_2024}

\subsection{Multiscale Template Matching}
In the process of multiscale template matching \cite{template_matching}, a set of templates representing various hand shapes, including thumbs up, scissors, paper, and rock, is utilized. These templates are carefully chosen to cover a diverse range of gestures commonly associated with hand interactions.

\begin{figure}[H]
\centering
\includegraphics[width=0.8\textwidth]{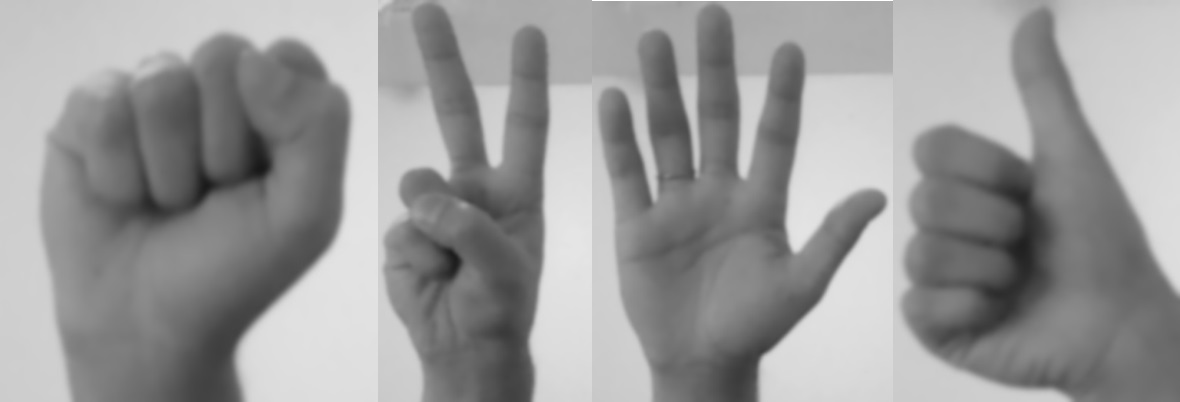}
\caption{Templates used for matching.}
\end{figure}

Let \( I \) be the input image, and \( T \) be a template image. The normalized cross-correlation \( R \) between the template \( T \) and a subregion of the input image \( I \) is calculated using the formula:

\begin{equation}\
R(u, v) = \frac{\sum_{x,y} (I(x, y) - \bar{I})(T(x-u, y-v) - \bar{T})}{\sqrt{\sum_{x,y} (I(x, y) - \bar{I})^2 \sum_{x,y} (T(x-u, y-v) - \bar{T})^2}} \
\end{equation}

\(R(u, v)\) represents the normalized cross-correlation between the template \(T\) and a subregion of the input image \(I\) at scale s, where \(u\) and \(v\) denote the top-left corner coordinates of the subregion. 
\begin{equation}
\text{Matching accuracy} = \max_{u, v, s} (R(u, v, s)) \
\end{equation}
This notation specifies that the maximum is taken over the variables \(u\), \(v\), and \(s\) and it represents the highest correlation value in the cross-correlation maps across all scales s of the image \(I\), indicating the best match between the template and the image subregions at the corresponding location. This notation indicates that the template \(T\) which maximizes the matching accuracy (\(MA_{T}\)) across all possible templates is selected as the best match with the image.
\[ \text{Predicted class} = \arg\max_{T} (MA_{T}) \]

The \textit{classify()} function is an integral part of the hand shape detection system, utilizing mathematical formulations to pinpoint the most suitable template among rock, paper, scissors, or thumbs up categories. The initial step involves converting the input image frame to grayscale, reducing it to a single channel for streamlined processing. Following this, the function employs a blurring operation on the template to optimize its features for efficient matching against the grayscale image. This blurring technique serves to diminish noise and enhance the robustness of the template matching process. In addition, the function adopts a multi-scale approach, performing template matching at various image scales ranging from 0.1 to 1. This strategy ensures adaptability to different hand sizes and distances from the camera, enhancing the versatility of the hand shape detection system. Once the most appropriate template is identified, the function displays the recognized template's name on the input frame. Alongside this, it provides a matching accuracy score associated with the selected template. This accuracy score quantifies the alignment between the identified hand shape and the chosen template, offering valuable insights into the reliability of the classification. An essential feature of the function is its feedback mechanism. If the highest matching accuracy falls below the empirically determined threshold of 0.74, the function reports "No Hand Detected." This threshold ensures that only confident matches are considered, contributing to the overall reliability of the hand shape detection system. For a detailed examination of the code implementation, you can refer to the source code available at the following GitHub repository: \href{https://github.com/Ketansuhaas/Hand-shape-detection}{https://github.com/Ketansuhaas/Hand-shape-detection}.

\subsection{Background Subtraction and Thresholding}

To isolate the hand region from the background, background subtraction \cite{bg_subtraction} is applied. The difference image (\texttt{diff}) is obtained by calculating the absolute difference between the background image \( B \) and the current frame \( F \). A binary threshold is then applied to \texttt{diff}, resulting in a binary image (\texttt{thresholded\_diff}), where the hand region is highlighted.

\begin{figure}[H]
\centering
\includegraphics[width=0.3\textwidth]{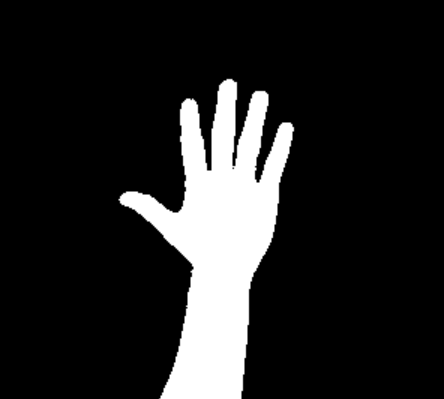}
\caption{An example of background subtracted and thresholded image.}
\end{figure}

The difference image \( \text{diff} \) is calculated as the absolute difference between the background image \( B \) and the current frame \( F \):\
\begin{equation}\
\text{diff}(x, y) = |B(x, y) - F(x, y)| \
\end{equation}

A binary threshold is applied to the difference image \( \text{diff} \) to obtain a binary image \( \text{thresholded\_diff} \):\

\begin{equation}\
\text{thresholded\_diff}(x, y) = \begin{cases} 255, & \text{if } \text{diff}(x, y) > \text{threshold} \\ 0, & \text{otherwise} \end{cases} \
\end{equation}

In the resulting binary image, the hand region is highlighted in white (pixel value 255), while the background appears in black (pixel value 0). This binary representation effectively isolates the hand from the background, providing a clear visual distinction for subsequent processing steps. In our implementation, we employed the initial frame as the background image, assuming that no other moving objects were present when it was captured. Additionally, we encountered an issue where the camera was initializing with a low resolution. To mitigate this, a delay of 2 seconds was introduced before capturing the first frame.

\subsection{Centroid Calculation}
We choose to use the largest contour under the assumption that the hand is the most substantial moving object within the frame. The segmentation of contours is accomplished through the application of the `cv2.findContours` function in OpenCV, which leverages the "chain code" or "contour tracing" algorithm. This algorithm systematically traces the boundaries of connected components within a binary image, effectively identifying distinct contours based on their pixel connectivity.
The centroid \( (C_x, C_y) \) of the largest contour is calculated using the formulae:\
\[
C_x = \frac{1}{M_{00}} \sum_x \sum_y x \cdot I_{\text{largest\_contour}}(x, y)
\]

\[
C_y = \frac{1}{M_{00}} \sum_x \sum_y y \cdot I_{\text{largest\_contour}}(x, y)
\]

And the zeroth moment \cite{binary_images} (\(M_{00}\)) would be calculated as:

\[
M_{00} = \sum_x \sum_y I_{\text{largest\_contour}}(x, y)
\]

\subsection{Bounding Boxes and Centroid Tracking}
The `\textit{draw\_ROI()}` function plays a pivotal role in our research implementation, specifically focusing on gesture detection and visualization of hand movements. This function begins by determining the extreme coordinates of the largest contour, a critical step that facilitates the creation of a bounding rectangle around the identified contour. This bounding rectangle serves as a visual representation of the hand shape in the input image frame. One of the prominent features of this function is the real-time plotting of the centroid on the image frame. The centroid, calculated from the identified contour, provides a dynamic point of reference, enhancing the interpretability of hand movements. To add a layer of sophistication to the gesture detection mechanism, the function maintains a record of the centroid in the previous frame. It actively monitors the horizontal difference between the current and previous centroids. If this difference exceeds 15 pixels, it signals hand movement, offering a reliable indication of dynamic gestures. Moreover, the function incorporates a safeguard mechanism. If the area of the largest contour is found to be less than 1000, a threshold determined empirically, the function displays "No Hand Detected." This ensures that only substantial hand shapes are considered for further analysis, contributing to the robustness of the gesture detection system. In terms of user feedback, when the system identifies a moving hand with a rock handshape, it outputs the message "Don't hit me with that rock!" For a paper handshape, the system displays the message "Bye!"—adding a touch of interactivity and engagement to the application. The integration of these mathematical formulations and feedback mechanisms in the `\textit{draw\_ROI()}` function enhances the overall functionality of the application, providing not only visual representation but also meaningful insights into hand gestures and motions. 

\begin{figure}[H]
\centering
\includegraphics[width=0.7\textwidth]{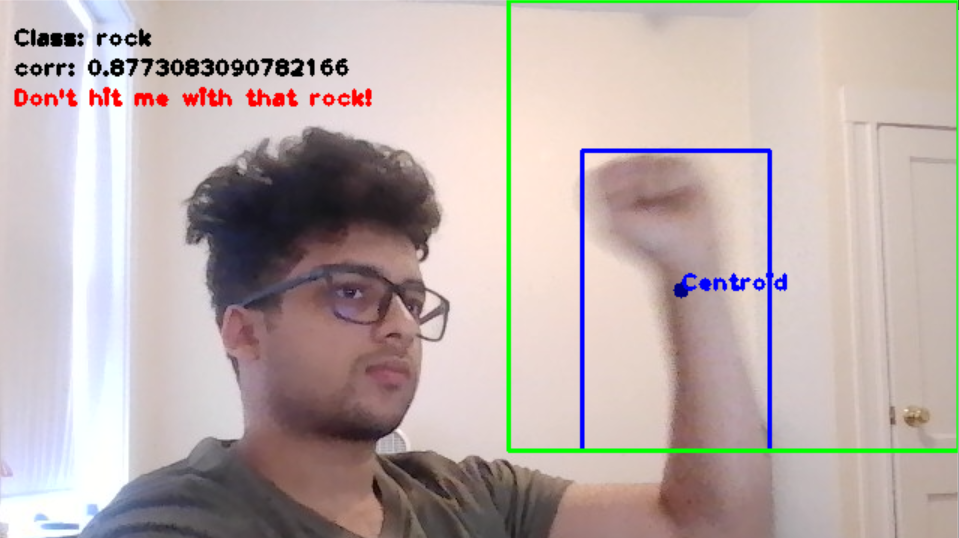}
\caption{Result of applying all the above Methods}
\end{figure}

\section{Experiments}
In this study, a comprehensive set of experiments was conducted to assess the robustness and performance of the proposed hand shape classification system under various conditions. The experiments aimed to evaluate the system's ability to classify hand shapes accurately in real-time scenarios. Key experimental factors included different lighting conditions, variations in hand distances from the camera, diverse hand positions, and varying hand orientations. These factors were systematically manipulated to simulate real-world scenarios and assess the system's performance across a range of challenging conditions. The experimentation involved a total of 3 tests, with each focused on assessing the system's robustness in specific areas: lighting conditions, hand distances, and positions. The tests were designed to thoroughly evaluate the system's performance under each factor. Lighting conditions were adjusted to simulate both well-lit and low-light environments, hand distances were varied to assess performance at different proximities, and diverse hand positions were explored to gauge the system's adaptability to varying user interactions. Correct classification is inferred when the class of the template with the highest matching accuracy aligns with the ground truth. To evaluate the system's performance, a suite of metrics including F1 score and Accuracy were utilized.

\begin{equation}
\text{Accuracy} = \frac{TP + TN}{TP + TN + FP + FN }
\end{equation}

\begin{equation}
\text{Precision} = \frac{TP}{TP + FP}
\end{equation}

\begin{equation}
\text{Recall} = \frac{TP}{TP + FN}
\end{equation}

\begin{equation}
\text{F1} = \frac{2 \cdot \text{Precision} \cdot \text{Recall}}{\text{Precision} + \text{Recall}}
\end{equation}

These metrics provide a comprehensive assessment of the classification accuracy, precision of positive classifications, ability to capture true positive instances, and overall correctness, contributing to a thorough evaluation of the system's efficacy.

\section{Results}
In this section, we present the results of our experimental evaluations designed to assess the effectiveness and resilience of the proposed hand shape classification system. Our study systematically investigates various factors that impact the real-time accuracy of the system, especially in dynamic scenarios. The findings are categorized into three primary areas, each addressing a distinct aspect of variability. The frames used for classification are extracted from video sequences, enabling a detailed examination of the continuity in classification. The initial phase involved a lighting variability experiment, where all four hand shapes were classified under specific lighting conditions. The lighting conditions were systematically altered for each set of trials, providing insights into the system's adaptability to varying illumination scenarios. Following the lighting variability experiment, we conducted the translational variability experiment, involving moving the hand within the same plane at a distance from the camera. This scenario allowed us to assess the system's accuracy in capturing hand shapes under various translational movements. Subsequently, we explored the proximity case, where the hand was moved forward and backward. This experiment aimed to assess the system's response to moving the hand along the axis through the camera or the depth axis. Each phase of the experiment contributes to a comprehensive understanding of the hand shape classification system's performance under diverse and challenging conditions. A demo video is available at \href{https://www.youtube.com/watch?v=QKJI6jGbKA4}{https://www.youtube.com/watch?v=QKJI6jGbKA4}.

\subsection{Lighting Variability}
This section presents the results of the experiments conducted to assess the robustness of the proposed hand shape classification system under diverse lighting conditions. The analysis delves into the system's performance when subjected to differential illumination, offering insights into its adaptability to varying lighting scenarios.

\begin{table}[H]
  \centering
  \begin{tabular}{|c|c|c|c|c|}
    \hline
    & \textbf{Predicted Rock} & \textbf{Predicted Thumbs Up} & \textbf{Predicted Scissors} & \textbf{Predicted Paper} \\
    \hline
    \textbf{Actual Rock} & 18 & 2 & 0 & 0 \\
    \hline
    \textbf{Actual Thumbs Up} & 1 & 18 & 1 & 0 \\
    \hline
    \textbf{Actual Scissors} & 3 & 0 & 16 & 1  \\
    \hline
    \textbf{Actual Paper} & 3 & 0 & 2 & 15   \\
    \hline
  \end{tabular}
  \caption{Confusion Matrix for Lighting Variability}
  \label{table:translational_confusion_matrix}
\end{table}

The majority of incorrect predictions can be attributed to insufficient lighting conditions, with the more challenging predictions often associated with hand gestures featuring intricate details. This phenomenon is particularly evident in the consistently higher accuracy of predicting "rock" compared to "paper."
\begin{figure}[H]
\centering
\includegraphics[width=0.9\textwidth]{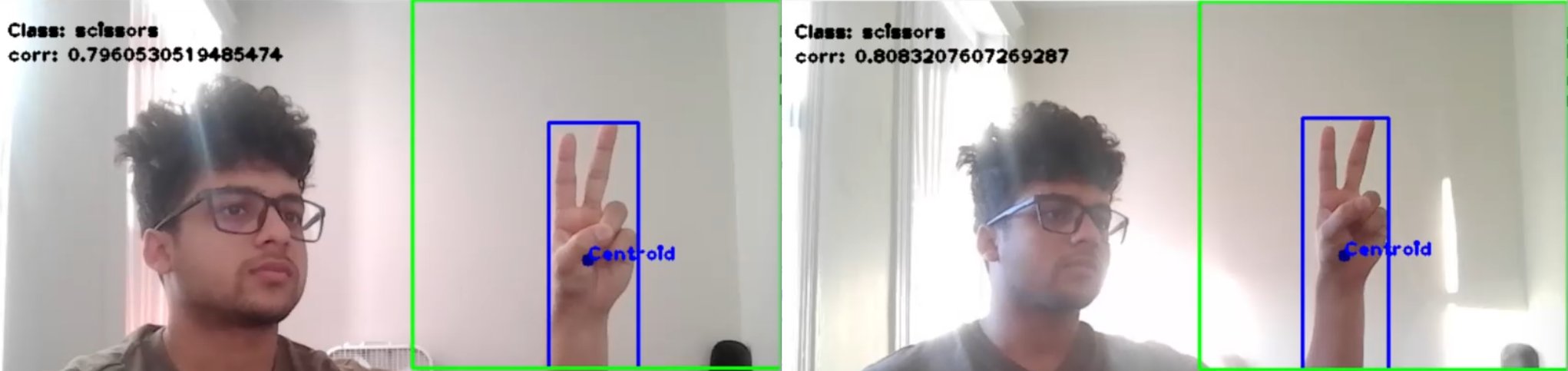}
\caption{Scissors classified at two different lighting conditions}
\end{figure}

\subsection{Translational Variability}
In this subsection, we examine the outcomes of experiments focused on translational variability. The assessment provides an in-depth understanding of how the system copes with changes in hand position along the same plane, shedding light on its performance under different translational conditions. 
\begin{table}[H]
  \centering
  \begin{tabular}{|c|c|c|c|c|}
    \hline
    & \textbf{Predicted Rock} & \textbf{Predicted Thumbs Up} & \textbf{Predicted Scissors} & \textbf{Predicted Paper} \\
    \hline
    \textbf{Actual Rock} & 19 & 1 & 0 & 0 \\
    \hline
    \textbf{Actual Thumbs Up} & 1 & 18 & 1 & 0 \\
    \hline
    \textbf{Actual Scissors} & 2 & 0 & 18 & 0 \\
    \hline
    \textbf{Actual Paper} & 1 & 0 & 1 & 18 \\
    \hline
  \end{tabular}
  \caption{Confusion Matrix for Translational Variability}
  \label{table:translational_confusion_matrix}
\end{table}

The inaccuracies in predictions are not solely attributable to straightforward translation but rather stem from subtle variations in orientation induced by human movements during continuous hand movements.

\begin{figure}[H]
\centering
\includegraphics[width=0.9\textwidth]{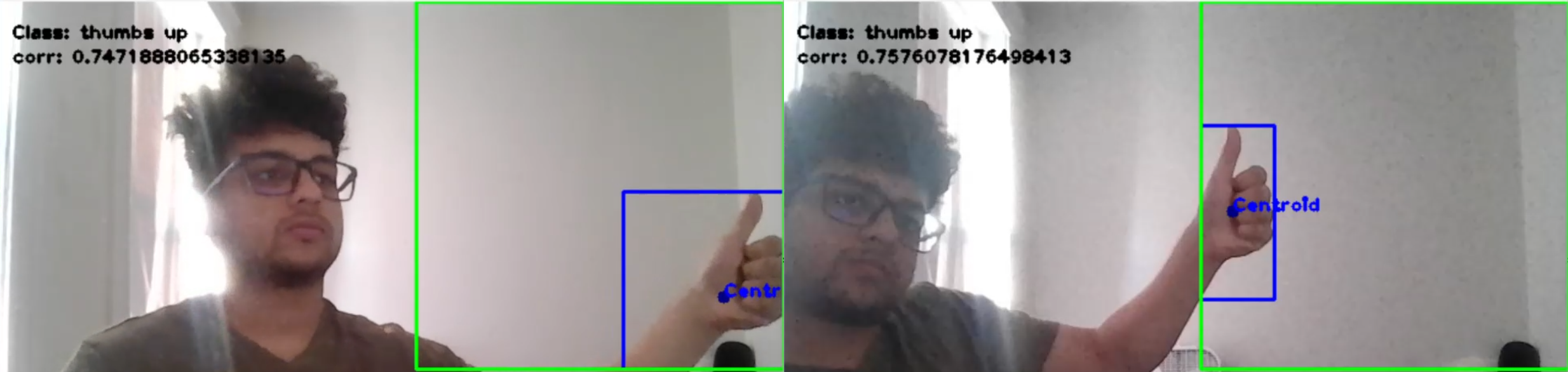}
\caption{Paper classified at two different positions}
\end{figure}

\subsection{Proximity Variability}
The results of experiments investigating proximity variability are detailed in this section. The evaluation explores the system's capability to accurately classify hand shapes when the distance between the hand and the camera varies. The findings offer valuable insights into the system's response to changes in proximity, providing a comprehensive understanding of its performance in different spatial configurations.
\begin{table}[H]
  \centering
  \begin{tabular}{|c|c|c|c|c|}
    \hline
    & \textbf{Predicted Rock} & \textbf{Predicted Thumbs Up} & \textbf{Predicted Scissors} & \textbf{Predicted Paper} \\
    \hline
    \textbf{Actual Rock} & 19 & 1 & 0 & 0 \\
    \hline
    \textbf{Actual Thumbs Up} & 0 & 19 & 1 & 0 \\
    \hline
    \textbf{Actual Scissors} & 2 & 0 & 18 & 0 \\
    \hline
    \textbf{Actual Paper} & 3 & 0 & 1 & 16 \\
    \hline
  \end{tabular}
  \caption{Confusion Matrix for Proximity Variability}
  \label{table:proximity_confusion_matrix}
\end{table}

In scenarios where the "paper" hand shape is situated at considerable distances during proximity variations, the challenges associated with capturing intricate details are heightened for both the template and the camera. This distance-related effect significantly amplifies the difficulties encountered in achieving accurate predictions. It is noteworthy that a comparable phenomenon is observed when the "scissors" hand shape is positioned at far distances. However, the impact is more pronounced in the case of "paper" due to its inherent intricacies. The complex details of the "paper" hand shape present additional hurdles in obtaining precise predictions, particularly when it is located at extended distances during proximity variations.
\begin{figure}[H]
\centering
\includegraphics[width=0.9\textwidth]{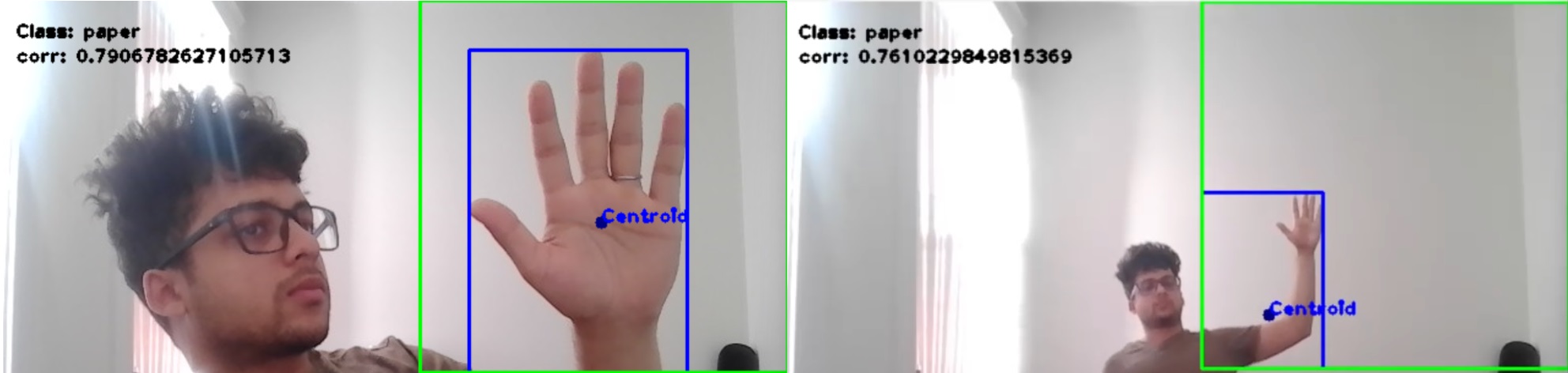}
\caption{Paper classified at two different proximities}
\end{figure}
\subsection{Comprehensive Evaluation}

This combined confusion matrix represents the total predictions made across all experiments. Each cell in the matrix indicates the number of instances where the predicted class aligns with the actual class.

\begin{table}[H]
  \centering
  \begin{tabular}{|c|c|c|c|c|}
    \hline
    & \textbf{Predicted Rock} & \textbf{Predicted Thumbs Up} & \textbf{Predicted Scissors} & \textbf{Predicted Paper} \\
    \hline
    \textbf{Actual Rock} & 56 & 4 & 0 & 0 \\
    \hline
    \textbf{Actual Thumbs Up} & 2 & 55 & 3 & 0 \\
    \hline
    \textbf{Actual Scissors} & 7 & 0 & 52 & 1 \\
    \hline
    \textbf{Actual Paper} & 7 & 0 & 4 & 49 \\
    \hline
  \end{tabular}
  \caption{Combined confusion matrix}
  \label{table:proximity_confusion_matrix}
\end{table}
Here is a comprehensive table showcasing the Average F1 scores and accuracies across different variability conditions. The table offers a nuanced understanding of how the system responds to diverse challenges posed by lighting variations, translational movements, and proximity changes. This collective assessment serves as a valuable resource for evaluating the system's robustness and identifying potential areas for refinement or enhancement.

\begin{table}[H]
  \centering
  \begin{tabular}{|c|c|c|c|c|c|c|}
    \hline
    \textbf{Variability} & \textbf{Accuracy} & \textbf{Average F1} \\
   \hline
    \textbf{Lighting} & 0.837 & 0.838 \\
    \hline
    \textbf{Translational} & 0.912 & 0.913 \\
    \hline
    \textbf{Proximity} & 0.9 & 0.9 \\
    \hline
    \textbf{Overall} & 0.881 & 0.882\\
    \hline
  \end{tabular}
  \caption{Performance in various conditions}
  \label{table:proximity_confusion_matrix}
\end{table}

The system's performance is notably compromised in low-light conditions, signaling a potential area for enhancement. To address this, a strategic approach involves refining the templates by incorporating more comprehensive representations derived from multiple photos. Alternatively, introducing templates tailored for diverse lighting conditions could contribute to a more robust performance. Proximity variations pose a challenge, particularly when the hand is positioned at a significant distance. This challenge is particularly pronounced with the paper hand shape due to the intricate details involved. Mitigating this issue might involve refining templates specifically for extended distances or exploring techniques to handle intricate hand shapes more effectively. Conversely, the system excels in translational scenarios, demonstrating its proficiency in accurately classifying hand shapes within the same plane. This success in translation suggests a robust capability to handle variations in hand positions along a consistent axis. These observations lay a foundation for targeted improvements, guiding the system toward enhanced performance across a spectrum of challenging conditions.

\section{Discussion}
\subsection{Strengths and Weaknesses of the Method}
The method exhibits notable strengths in its adept handling of translational and proximity variations, showcasing robust adaptability to shifts in hand position and changes in distance from the camera. However, its susceptibility to lighting conditions emerges as a weakness, resulting in a decrease in accuracy under varying illumination. Moreover, the method faces challenges in low-lighting conditions, where reduced performance is observed. It also struggles when the hand is positioned very far away, especially in proximity conditions, notably for the paper and scissors hand shapes. The intricate details associated with these shapes contribute to the difficulties encountered in accurate detection under such circumstances. Changes in background illumination after fixing the background pose an additional challenge, distorting the background-subtracted image, introducing noise, and making it difficult to detect the hand contour. The method also grapples with hand shape detection when subjected to rotations due to the lack of rotational homogeneity in its fixed-orientation templates. Another constraint is the requirement for a static background, limiting its application to scenarios where the hand is the predominant moving element within the frame.

\subsection{Comparison with Expectations}
Contrary to expectations, the method surpassed anticipations, delivering commendable performance even when challenged with hastily captured templates in suboptimal lighting conditions. This exceeded initial expectations and showcased resilience, particularly when compared to the often more complex and resource-intensive deep learning approaches. The method demonstrated a notable blend of efficiency and speed in classifying hand shapes under various conditions.

\subsection{Potential Future Work}
\begin{enumerate}
    \item \textbf{Rotational Homogeneity:} Future iterations should focus on expanding the template dataset to encompass multiple orientations of hand shapes, rectifying the current limitation associated with rotational homogeneity \cite{rotation}. This step would contribute to enhancing the system's overall adaptability and performance.
    
    \item \textbf{Dynamic Background Handling:} Addressing the static background requirement is crucial. Future efforts could explore strategies to make the system adaptable to dynamic backgrounds, thereby increasing its relevance and applicability in real-world environments \cite{robust}.
    
    \item \textbf{Incorporating Temporal Information:} Considering the system's performance in dynamic scenarios, incorporating temporal information holds promise. This enhancement could lead to more accurate and context-aware hand shape classification, especially in situations involving dynamic gestures.
    
    \item \textbf{Template Quality Enhancement:} Prioritizing template quality improvement is essential, given the method's vulnerability in low lighting conditions. Future work should delve into advanced template creation techniques or preprocessing methods to bolster template robustness, ensuring consistent accuracy even in challenging scenarios.
\end{enumerate}

By strategically addressing these aspects in future iterations, the method can fortify its weaknesses and enhance its versatility and reliability across a broader spectrum of real-world scenarios.\\

\section{Conclusions}
In conclusion, the presented hand shape classification approach utilizing multiscale template matching demonstrates commendable performance in basic hand shape classification tasks. Despite the method's simplicity, it excels in handling translational and proximity variations, showcasing robust adaptability to shifts in hand position and changes in distance from the camera. The experimental results highlight its effectiveness in controlled environments, providing a foundation for potential applications in straightforward human-computer interaction scenarios.

One of the notable strengths of the method lies in its adaptability to translational variations, allowing it to accurately classify hand shapes within consistent axes. This characteristic aligns well with real-world scenarios where users may not maintain fixed hand positions during interactions. Additionally, the method's proficiency in handling proximity variations underscores its versatility, making it well-suited for applications where users engage with the system from different distances.

However, the method faces challenges in dealing with varying lighting conditions and specific hand positions, particularly for intricate hand shapes like paper and scissors. Reduced performance in low-lighting conditions and difficulties in accurate detection when the hand is placed very far away highlight areas for improvement. Despite these challenges, the method has exceeded expectations by delivering robust performance even in suboptimal lighting conditions.

The method's sensitivity to changes in background illumination and the lack of rotational homogeneity in fixed-orientation templates are identified as limitations. Nevertheless, its efficiency and speed in classifying hand shapes under various conditions position it as a practical alternative, especially when compared to more complex deep learning approaches, particularly in scenarios with limited data.

Importantly, it is worth noting that with additional refinement and addressing the identified limitations, the method could be on par with or even surpass deep learning techniques in certain scenarios. Given more time and resources, the method has the potential to achieve comparable accuracy levels to deep learning methods while maintaining its efficiency and simplicity. This underscores the method's promise as a competitive alternative for real-time hand shape classification, offering a valuable option for applications where deep learning techniques may face challenges or resource constraints.

\bibliographystyle{plain}  
\bibliography{references}  

\begin{thebibliography}{1}

\bibitem{binary_images}
Binary images.
\newblock \url{https://homepages.inf.ed.ac.uk/rbf/CVonline/LOCAL_COPIES/OWENS/LECT2/node3.html}.
\newblock Accessed on February 11, 2024.

\bibitem{OpenCV_2024}
{OpenCV - Open Computer Vision Library}.
\newblock \url{https://opencv.org/}.
\newblock Accessed on February 11, 2024.

\bibitem{template_matching}
Template matching.
\newblock \url{https://en.wikipedia.org/wiki/Template_matching#:~:text=Template%20matching%20is%20a%20technique,or%20edge%20detection%20in%20images.}
\newblock Accessed on February 11, 2024.

\bibitem{robust}
Frederic Jurie and Michel Dhome.
\newblock Real time robust template matching.
\newblock In {\em Proceedings of the British Machine Vision Conference 2002}, 09 2002.

\bibitem{rotation}
Hae Kim and Sidnei Araújo.
\newblock Grayscale template-matching invariant to rotation, scale, translation, brightness and contrast.
\newblock pages 100--113, 12 2007.

\bibitem{bg_subtraction}
M.~Piccardi.
\newblock Background subtraction techniques: a review.
\newblock In {\em 2004 IEEE International Conference on Systems, Man and Cybernetics (IEEE Cat. No.04CH37583)}, volume~4, pages 3099--3104 vol.4, 2004.

\end{thebibliography}

\end{document}